\documentclass[letterpaper, 10 pt, conference]{ieeeconf}  
\usepackage{url}
\usepackage{graphicx}
\usepackage{amsmath}
\usepackage{multirow}
\usepackage{caption}
\usepackage{subcaption}
\usepackage{soul}
\usepackage[usenames,dvipsnames]{color}

\IEEEoverridecommandlockouts                              

\title{\LARGE \bf
SurgPose: Generalisable Surgical Instrument Pose Estimation using Zero-Shot Learning and Stereo Vision
}

\author{Utsav Rai$^{1}$, Haozheng Xu$^{2}$ and Stamatia Giannarou$^{2}$
\thanks{*Stamatia Giannarou and Haozheng Xu are supported by the Royal Society [URF$\setminus$R$\setminus$201014].}
\thanks{$^{1}$Utsav Rai is with the Department of Computing, Imperial College London, UK
        {\tt\small utsav.rai23@imperial.ac.uk}}%
\thanks{$^{2}$Haozheng Xu and Stamatia Giannarou are with the Hamlyn Centre for Robotic Surgery, Department of Surgery and Cancer, Imperial College London, UK
        {\tt\small {haozheng.xu19, stamatia.giannarou}@imperial.ac.uk}}%
}

\begin{document}

\maketitle
\thispagestyle{empty}
\pagestyle{empty}

\begin{abstract}

Accurate pose estimation of surgical tools in Robot-assisted Minimally Invasive Surgery (RMIS) is essential for surgical navigation and robot control. While traditional marker-based methods offer accuracy, they face challenges with occlusions, reflections, and tool-specific designs. Similarly, supervised learning methods require extensive training on annotated datasets, limiting their adaptability to new tools. Despite their success in other domains, zero-shot pose estimation models remain unexplored in RMIS for pose estimation of surgical instruments, creating a gap in generalising to unseen surgical tools. This paper presents a novel 6 Degrees of Freedom (DoF) pose estimation pipeline for surgical instruments, leveraging state-of-the-art zero-shot RGB-D models like the FoundationPose and SAM-6D. We advanced these models by incorporating vision-based depth estimation using the RAFT-Stereo method, for robust depth estimation in reflective and textureless environments.
Additionally, we enhanced SAM-6D by replacing its instance segmentation module, Segment Anything Model (SAM), with a fine-tuned Mask R-CNN, significantly boosting segmentation accuracy in occluded and complex conditions. Extensive validation reveals that our enhanced SAM-6D surpasses FoundationPose in zero-shot pose estimation of unseen surgical instruments, setting a new benchmark for zero-shot RGB-D pose estimation in RMIS. This work enhances the generalisability of pose estimation for unseen objects and pioneers the application of RGB-D zero-shot methods in RMIS.

\end{abstract}

\section{INTRODUCTION}
Accurate real-time pose estimation of surgical tools is a critical enabler of Robot-assisted Minimally Invasive Surgery (RMIS), ensuring precision, safety, and optimal control during delicate procedures. Existing pose estimation techniques in surgical robotics primarily rely on either marker-based methods or supervised learning approaches, both of which pose significant challenges in the surgical environment. Marker-based methods \cite{cartucho2022enhanced}, \cite{wang2022towards} \cite{super} though widely adopted, require specialised markers designed for sterile surgical settings. These markers are prone to occlusion, interference from reflective surfaces, and require precise sensor fusion, often depending on the robot’s kinematic \cite{wang2022towards}, \cite{hao2018vision}, \cite{lu2021superdeep} data to estimate the tool’s pose. However, the kinematic measurements, especially in cable-driven robotic systems, frequently lack calibration, resulting in inaccuracies. Methods such as \cite{wang2022towards} propose automatic robot calibration to address these issues, but are constrained by hardware limitations. In addition, stereo vision has been integrated into some marker-based systems to incorporate depth information \cite{wang2020image}, \cite{hao2018vision}, but they rely on conventional computer vision algorithms for depth estimation, which struggle in the complex environments typical of surgery. These methods also necessitate specific marker designs for new tools, and fail to robustly handle occlusions or highly reflective surfaces.

Supervised learning-based methods have emerged to overcome many of the limitations associated with marker-based approaches \cite{xu2023graph}. \cite{xu2023graph} offer markerless pose estimation by directly learning the 2D-3D correspondences between the RGB images and 3D CAD models of the tools. Such approaches have demonstrated robustness in scenarios involving occlusion, and they eliminate the need for physical markers. However, supervised learning comes with its own challenges—each new tool requires a large, annotated dataset to achieve reliable accuracy. Consequently, these methods are not generalisable to unseen tools without retraining, which limits their practicality in dynamic surgical environments where new instruments are frequently introduced.

Zero-shot pose estimation methods represent a promising alternative, offering the ability to generalise to new, unseen objects without retraining \cite{wen2023foundationpose},\cite{lin2024sam6dsegmentmodelmeets},\cite{cai2022ove6d},\cite{labbe2022megapose}.
However, these methods typically rely on depth information captured by depth sensors, which is impractical for surgical applications due to the close-range environment and limitations such as sensor blind zones and interference from reflective surfaces. While these methods have shown strong zero-shot transfer capabilities across a variety of domains, they have not yet been applied in the context of RMIS. The challenge in applying zero-shot pose estimation for surgical instruments lies in the complexity of accurately segmenting the tool and obtaining reliable depth information. A robust zero-shot pose estimation pipeline for RMIS would require accurate inputs, including a mask of the visible tool, a depth image, an RGB image, and a 3D CAD model of the tool.

One of the major hurdles in RMIS is the lack of reliable depth information. 
While monocular depth estimation techniques have been explored \cite{yang2024depth}, \cite{piccinelli2024unidepth}, \cite{hu2024metric3d}, they frequently suffer from inconsistent scale and depth inaccuracies, which negatively impact pose estimation performance. Stereo vision offers a promising alternative for depth estimation, but traditional stereo algorithms struggle in reflective, textureless, and occluded environments typical of surgery \cite{Geiger2010ACCV}. Methods like \cite{super}, \cite{lu2021superdeep} have integrated depth from stereo. However, \cite{super} relies on conventional computer vision techniques like \cite{Geiger2010ACCV}, which are prone to inaccuracies due to high reflections from metallic tools, while \cite{lu2021superdeep} requires large-scale ground truth datasets for training deep neural networks to accurately calculate disparity. Advanced disparity estimation methods like \cite{lipson2021raft} have demonstrated significantly improved performance in challenging environments, with zero-shot capabilities that generalise effectively to surgical settings, offering a more robust and viable solution for surgical applications.

In this paper, we bridge the gap by proposing a novel, end-to-end zero-shot pose estimation pipeline tailored for RMIS. Our pipeline leverages state-of-the-art zero-shot RGB-D models like the FoundationPose and SAM-6D. 
The key contributions of this work are as follows: 
\begin{itemize} 
\item We advanced RGB-D pose estimation models by incorporating vision-based depth estimation using the RAFT-Stereo method instead of relying on depth sensors. This enables robust depth estimation in reflective and textureless environments and expands the applicability of our method to different scenarios.
\item We enhanced the SAM-6D zero-shot pose estimation model by integrating a fine-tuned Mask R-CNN \cite{he2018maskrcnn}, leading to improved segmentation accuracy in occluded and reflective surgical environments. 
\item We introduced a novel database for surgical instrument pose estimation, which includes both real and synthetic stereo images with ground truth tool pose information, allowing validation of pose estimation methods.
\item We conducted a comprehensive evaluation of state-of-the-art zero-shot pose estimation models in RMIS, highlighting their strengths, limitations, and the improvements gained from our proposed enhancements. \end{itemize}

By addressing the unique challenges posed by surgical environments—such as occlusions, reflective surfaces, and the absence of reliable depth sensing—this work provides a more robust and adaptable solution for RMIS. Our pipeline demonstrates the viability of zero-shot pose estimation in surgical robotics, setting a new standard for generalisability and accuracy in surgical instrument tracking and manipulation.

\begin{figure*}[t]
    \centering
    \includegraphics[width=0.75\textwidth]{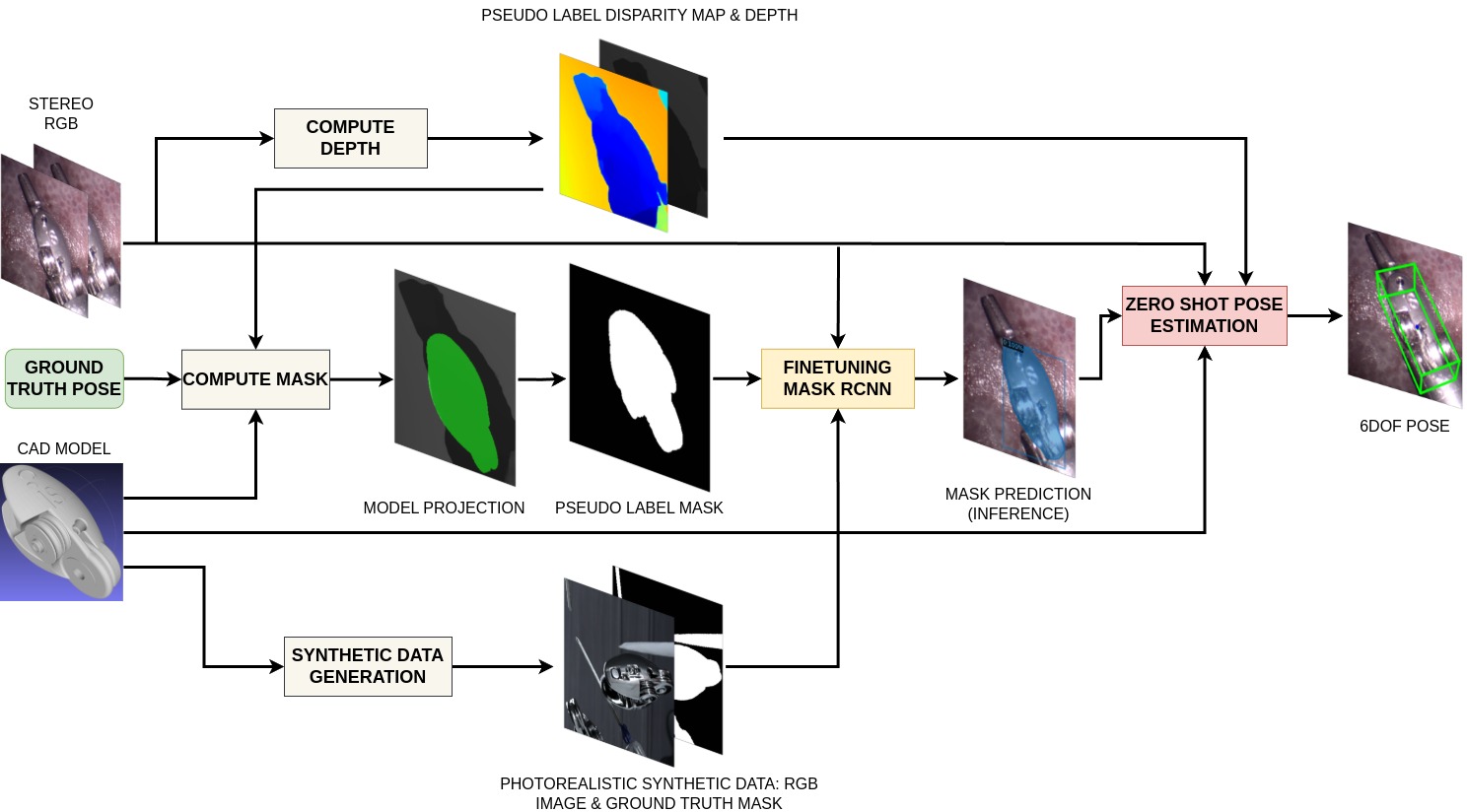}
    \captionsetup{belowskip=-5pt}
    \caption{Proposed pipeline for zero-shot pose estimation of surgical tools in RMIS.}
    \label{fig:pipeline}
    \vspace{-15pt}
\end{figure*}

\section{RELATED WORK}

Recent advances in 6D pose estimation have primarily focused on RGB and RGB-D data, leveraging deep learning techniques to handle complex environments. However, despite significant progress, these methods face limitations, particularly in RMIS, where tool occlusion, reflective surfaces, and limited data availability pose unique challenges. 

\subsection{RGB-Based Approaches}


\textbf{Indirect methods} establish 2D-3D correspondences, typically using a RANSAC based Perspective-n-Point (PnP) algorithm to estimate object pose. For instance, \cite{zakharov2019dpod}, \cite{park2019pix2pose}, \cite{li2019cdpn}, \cite{haugaard2022surfemb} and \cite{peng2019pvnet} detect 2D keypoints and match them to the object’s 3D model. Xu et al. \cite{xu2023graph} introduced a keypoint-graph-based network for RMIS applications, leveraging Graph Convolutional Networks (GCNs) to combine spatial and temporal data. This approach addresses challenges specific to surgical settings, such as partial visibility, by detecting 2D keypoints, refining them through a graph-based approach, and calculating the tool pose with a PnP solver. Although these methods can handle occlusion effectively and achieve high accuracy, they rely on iterative, non-differentiable solvers that limit their direct use in tasks requiring end-to-end learning or real-time adaptability.

\textbf{Direct methods}: While indirect methods, which rely on 2D-3D correspondences, often show strong performance, they are not suitable for tasks that require a fully differentiable process \cite{wang2020self6d}. In such cases, direct methods are employed, where the 6D pose is directly regressed from the input \cite{xiang2017posecnn}, \cite{labbe2020cosypose} and \cite{do2018lienet}. These approaches use point matching loss functions to minimise the distance between predicted and ground truth poses \cite{xiang2017posecnn}, \cite{labbe2020cosypose}, or alternatively, employ distinct loss functions for the rotation and translation components \cite{do2018lienet}. \cite{wang2021gdrnetgeometryguideddirectregression}, in particular, enhances accuracy by learning geometric guidance from dense correspondence-based intermediate representations. While these methods demonstrate impressive results in cluttered and occluded environments, they rely on large-scale, annotated datasets, which are difficult to obtain in RMIS due to privacy concerns and the need for constant retraining as new tools are introduced.

\subsection{RGB-D-Based Approaches}

RGB-D methods incorporate depth data alongside RGB images, offering improved pose estimation accuracy, especially in complex and occluded environments. Methods like \cite{he2020pvn3d} and \cite{he2021ffb6d} establish 3D-3D correspondences between the object’s 3D model and the depth data, significantly improving performance in challenging scenarios. However, these methods require accurate ground truth depth during training and inference, which is difficult to obtain in RMIS due to the limitations of depth sensors in surgical settings, such as reflections and blind zones in close proximity. Importantly, RGB-D-based deep learning approaches have not yet been explored in RMIS, highlighting a key gap in the current research landscape.

Zero-shot pose estimation models, such as FoundationPose \cite{wen2023foundationpose}, SAM-6D \cite{lin2024sam6dsegmentmodelmeets}, \cite{labbe2022megapose} and OVE-6D \cite{cai2022ove6d}, also rely on depth information but have the distinct advantage of generalising to unseen objects without object-specific training. While RGB-D methods require accurate ground truth depth, zero-shot methods like \cite{wen2023foundationpose} and our proposed enhancement of SAM-6D \cite{lin2024sam6dsegmentmodelmeets}, as demonstrated in this work, perform well with pseudo-labelled depth estimated from stereo vision. These zero-shot transfer capabilities make these methods attractive for RMIS, though both RGB-D and zero-shot approaches remain largely unexplored in this domain.

This work addresses these gaps by evaluating state-of-the-art zero-shot models in RMIS and proposing enhancements to improve their performance in challenging surgical environments, specifically by leveraging stereo-based depth estimation instead of traditional depth sensors.

\section{Methodology}

This section presents the proposed pipeline for zero-shot 6DOF pose estimation of surgical tools, specifically designed for RMIS. 


\subsection{Pipeline Overview}
The proposed pipeline, illustrated in Figure \ref{fig:pipeline}, leverages stereo vision to estimate depth and generates a mask for the surgical tool, which is then used in zero-shot pose estimation models. 
The pipeline generates depth images from stereo images by calculating disparity using the RAFT-Stereo algorithm, which generalises well in RMIS environments without requiring further fine-tuning. For mask generation, a fine-tuned Mask R-CNN model is employed, as zero-shot segmentation methods tend to perform poorly in surgical environments due to the presence of occlusion and reflective surfaces. The following sections explain each stage of the pipeline in detail.

\subsection{Stereo-Based Depth Estimation}
The proposed method employs a stereo vision system and the deep learning-based disparity estimation technique, RAFT-Stereo, to calculate depth from stereo image pairs. RAFT-Stereo is selected due to its robustness in handling reflective and textureless environments commonly encountered in RMIS, outperforming conventional methods in these challenging scenarios.

\subsubsection{Disparity Calculation using RAFT-Stereo}
Disparity maps are generated from rectified stereo images using RAFT-Stereo, which refines disparity estimates iteratively through convolutional GRUs. RAFT-Stereo constructs a 3D correlation volume that encodes pixel similarities between the stereo image pairs. To pass information across the image, RAFT-Stereo uses multi-level GRU units. This iterative process ensures accurate disparity even in complex environments with reflective and textureless surfaces.


\subsubsection{Depth Calculation from Disparity}
The depth $Z(x, y)$ at each pixel is calculated from the disparity using the stereo baseline $B$ and the camera focal length $f$. The baseline $B$ represents the known distance between the two stereo cameras, determined during calibration. The focal length $f$ is derived from the intrinsic camera parameters. The computed depth map is critical for 6DOF pose estimation and is utilised in subsequent stages of the pipeline.

\subsection{Mask Generation Using Mask R-CNN}
The second stage of the pipeline focuses on generating accurate masks of the surgical tool using a fine-tuned Mask R-CNN model. Mask R-CNN is a supervised learning model trained on RGB images and corresponding pseudo label masks, providing robust segmentation even in challenging environments, such as those with occlusions, reflective surfaces, or limited visibility.

\subsubsection{Pseudo Label Mask Generation}
The pseudo label tool segmentation mask is generated by projecting the 3D CAD model of the surgical tool into the 2D image plane using the known ground truth tool pose and intrinsic camera parameters. The projection provides an initial segmentation mask, which is then refined by comparing the depth of the projected model, $Z_{\text{proj}}(u, v)$, with the depth obtained from the RAFT-Stereo disparity map, $Z_{\text{disp}}(u, v)$. A point is retained in the mask if:
\begin{equation}
|Z_{\text{proj}}(u, v) - Z_{\text{disp}}(u, v)| < \epsilon
\end{equation}

where, $\epsilon$ is a threshold set equal to 1 mm. The process defined in Eq. (1) helps to eliminate occluded regions from the mask, preserving only the visible cross-section of the tool. This method of refining the tool mask is crucial for improving segmentation accuracy, particularly in complex surgical environments where occlusions and reflections are common.

\subsubsection{Fine-tuning Mask R-CNN}
Mask R-CNN can be fine-tuned using different types of segmentation data namely, synthetic data, real data, or a mixture of both. Synthetic data generation offers a more automated and scalable approach, providing greater control over environmental conditions and enabling the generation of photorealistic training images with complex reflections, occlusions, and textures. 

Image rendering tools such as \cite{Denninger2023} and \cite{morrical2021nvisii} facilitate the creation of highly realistic synthetic datasets, with photorealistic renderings. Additionally, these tools allow the simulation of multiple instruments, distractors (objects not of interest in the scene), and environmental nuances, which helps to produce rich and diverse datasets that capture a variety of challenges, such as occlusions, truncation (partial visibility of objects due to being out of the camera’s field of view), and high reflections. By leveraging these techniques, ground truth instrument annotations can be automatically generated, eliminating the need for manual dataset capturing.

For a more automated and scalable pipeline, we fine-tuned Mask R-CNN using a combination of synthetic data and real images. This simplifies the data collection process while allowing the network to generalise better to realistic RMIS scenarios. Additionally, the depth maps obtained from RAFT-Stereo were used to calculate pseudo-label visible masks of surgical instruments, further improving segmentation accuracy by capturing the nuances of real surgical environments. This hybrid use of synthetic and real data enables the model to handle complex, occluded settings more effectively.

\subsubsection{Mask Generation During Inference}
During inference, the fine-tuned Mask R-CNN generates the segmentation mask from RGB images. The accuracy of the Mask R-CNN model is significantly improved by training it on a combination of synthetic and real data, as well as refining the ground truth mask using the depth-based occlusion handling defined in Eq. (1). This approach yields better segmentation performance than zero-shot segmentation models like SAM in RMIS environment.

\subsection{Zero-Shot RGB-D Pose Estimation}
The final stage of the proposed pipeline focuses on zero-shot RGB-D pose estimation. 
Zero-shot pose estimation models like FoundationPose \cite{wen2023foundationpose}, SAM-6D \cite{lin2024sam6dsegmentmodelmeets}, OVE6D \cite{cai2022ove6d}, and MegaPose \cite{labbe2022megapose} follow a general framework, utilising inputs such as RGB-D images, object mask, and CAD models to estimate the pose of objects without requiring specific training on those objects. 
These models process input data, such as RGB, RGB-D, or depth images combined with a segmentation mask, to detect and segment the surgical tool. For example, SAM-6D leverages the Segment Anything Model (SAM) \cite{kirillov2023segment} for segmentation, while the FoundationPose integrates foundation models, both relying on segmentation as a critical step for pose estimation.

For feature extraction, OVE6D primarily uses depth images and a segmentation mask, relying on a viewpoint encoder to capture the object’s orientation. MegaPose generates synthetic views of the CAD model and compares them with observed images to refine object pose. SAM-6D combines SAM with a 6D object pose estimator, enhancing segmentation robustness, while FoundationPose uses large-scale pre-trained models to generalise across diverse objects without retraining.

Once the features are extracted, an initial pose hypothesis is generated by aligning the object’s 2D image points with the 3D structure from the CAD model. For example, MegaPose uses a render-and-compare approach, while OVE6D leverages viewpoint codebooks and iterative in-plane rotation refinement. SAM-6D and FoundationPose refine the pose by optimising the alignment between the predicted pose and observed features, ensuring accurate object pose estimation.

In our proposed pipeline, SAM, the original instance segmentation module in SAM-6D, is replaced with the proposed fine-tuned Mask R-CNN. This fine-tuned Mask R-CNN was chosen as the segmentation model for all methods, including OVE6D, MegaPose and FoundationPose.


\section{Experiments and Results}
\subsection{Dataset Collection}

\begin{figure}[t]
    \centering
    \begin{subfigure}[t]{0.25\linewidth}
        \centering
        \includegraphics[width=\linewidth]{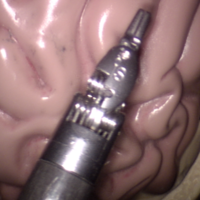}
        \caption{}
    \end{subfigure}
    \hspace{-7pt}
    \begin{subfigure}[t]{0.25\linewidth}
        \centering
        \includegraphics[width=\linewidth]{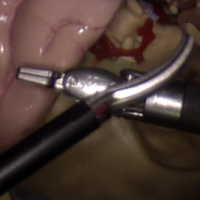}
        \caption{}
    \end{subfigure}
    \hspace{-7pt}
    \begin{subfigure}[t]{0.25\linewidth}
        \centering
        \includegraphics[width=\linewidth]{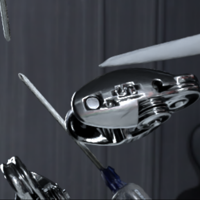}
        \caption{}
    \end{subfigure}
    \hspace{-7pt}
    \begin{subfigure}[t]{0.25\linewidth}
        \centering
        \includegraphics[width=\linewidth]{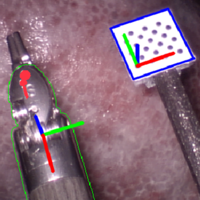}
        \caption{}
    \end{subfigure}
    \captionsetup{belowskip=-10pt}
    \caption{(a) Dataset A, (b) Dataset B, (c) Dataset C containing both synthetic and real images, and (d) Ground truth tool pose obtained using marker attachment.}
    \label{fig:gtpose}
    \vspace{-10pt}
\end{figure}

The primary goal of this study is to estimate the 6DOF pose of the Endowrist\texttrademark Large Needle Driver (LND), a commonly used surgical tool in the Da Vinci\texttrademark Si robotic system. For the purposes of this study, the pose estimation focused exclusively on the movable end joint, excluding the shaft. A ground truth pose was captured by attaching a Keydot marker \cite{opencv_library} to the shaft as shown in Fig. \ref{fig:gtpose} (d). Using stereo images, both pseudo label depth maps and visible masks were generated. Two datasets have been collected inside a simulated surgical environment created using organ phantoms, while a third dataset was created using a combination of real and synthetic images.

\textbf{Dataset A} consists of images of the non-occluded LND tool and contains 1027 images (Fig. \ref{fig:gtpose} (a)). \textbf{Dataset B} introduces instrument occlusions, using various surgical tools like laparoscopic instruments, surgical scissors and forceps to partially or fully obscure the LND tool (Fig. \ref{fig:gtpose} (b)) and contains 797 images.  Both datasets were collected under varying illumination conditions, naturally creating reflections on the tool surfaces. These datasets were used to evaluate the performance of our zero-shot pose estimation pipeline in non-occluded and occluded scenarios. Additionally, \textbf{Dataset C} has been created for fine-tuning the Mask R-CNN model. It comprised 1,489 images captured in both occluded and non-occluded scenarios, supplemented by 1,000 photorealistic synthetic images generated using the NVISII renderer \cite{morrical2021nvisii} (Fig. \ref{fig:gtpose} (c)).

\subsection{Hardware Setup}

The Da Vinci stereo camera with an original resolution of 1920x1080 per camera was utilised for stereo image capture, and the captured images were downsampled to 960x540. A 3D-printed attachment was used to mount a marker on the LND tool shaft (Fig. \ref{fig:gtpose} (d)) . The transformation between the marker and the LND tool was estimated using the approach mentioned in \cite{xu2023graph}.

\subsection{Performance Evaluation Metrics}


For validating the tool segmentation, we employed Average Precision (AP) and Average Recall (AR) at various Intersection over Union (IoU) thresholds. The AP@[IoU=0.50:0.95] metric provides the mean precision over IoU thresholds ranging from 0.50 to 0.95, capturing the model's performance across varying degrees of overlap between the predicted and ground truth masks. AR@[IoU=0.50:0.95] assesses the recall at these thresholds, providing insight into the model’s ability to detect objects consistently. Additionally, AP was computed for different object scales: AP$_s$ for small, AP$_m$ for medium, and AP$_l$ for large objects, to understand how well the model generalises to objects of different sizes.

To validate pose estimation, we used the Average Distance of Model Points (ADD) and the 2D Projection metrics. The ADD metric measures the average distance between corresponding points on the 3D models transformed by the predicted and ground truth poses. A pose is considered correct if this distance is below a given threshold, typically 1mm, 2.5mm, or 5mm. The 2D Projection metric computes the mean distance between the projections of 3D model points in the image plane, comparing the ground truth and predicted poses. If the distance is less than a set pixel threshold (e.g., 5px, 20px, or 50px), the pose is deemed correct. These metrics provide a robust evaluation of pose estimation accuracy, covering both 3D alignment and visual consistency in the 2D image.

In addition, we report the mean and standard deviation of both the ADD and 2D Projection metrics. These statistics provide a comprehensive view of the consistency and reliability of the pose estimation across different scenarios, helping to identify not only the average performance but also the variation in model predictions.

\subsection{Validation of Instrument Segmentation}

\begin{figure*}[ht]
    \centering
    \begin{subfigure}[b]{0.14\linewidth}
        \centering
        \includegraphics[width=\linewidth]{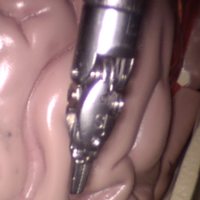}  
    \end{subfigure}
    \hspace{-7pt}
    \begin{subfigure}[b]{0.14\linewidth}
        \centering
        \includegraphics[width=\linewidth]{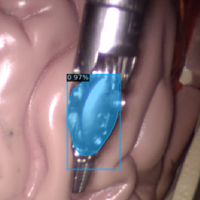}  
    \end{subfigure}
    \hspace{-7pt}
    \begin{subfigure}[b]{0.14\linewidth}
        \centering
        \includegraphics[width=\linewidth]{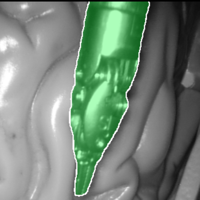}  
    \end{subfigure}
    \hspace{-7pt}
    \begin{subfigure}[b]{0.14\linewidth}
        \centering
        \includegraphics[width=\linewidth]{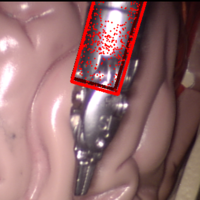}  
    \end{subfigure}
    \hspace{-7pt}
    \begin{subfigure}[b]{0.14\linewidth}
        \centering
        \includegraphics[width=\linewidth]{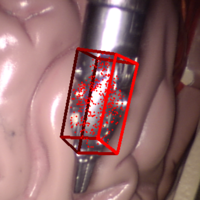}  
    \end{subfigure}
    \hspace{-7pt}
    \begin{subfigure}[b]{0.14\linewidth}
        \centering
        \includegraphics[width=\linewidth]{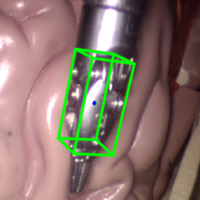}  
    \end{subfigure}

    \begin{subfigure}[b]{0.14\linewidth}
        \centering
        \includegraphics[width=\linewidth]{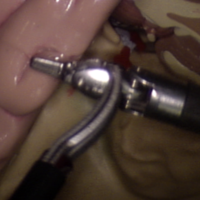}  
        \caption{}
    \end{subfigure}
    \hspace{-7pt}
    \begin{subfigure}[b]{0.14\linewidth}
        \centering
        \includegraphics[width=\linewidth]{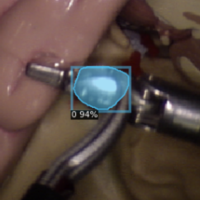}  
        \caption{}
    \end{subfigure}
    \hspace{-7pt}
    \begin{subfigure}[b]{0.14\linewidth}
        \centering
        \includegraphics[width=\linewidth]{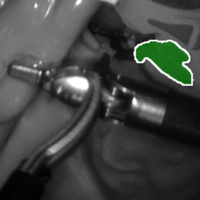}  
        \caption{}
    \end{subfigure}
    \hspace{-7pt}
    \begin{subfigure}[b]{0.14\linewidth}
        \centering
        \includegraphics[width=\linewidth]{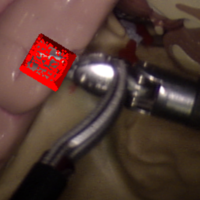}  
        \caption{}
    \end{subfigure}
    \hspace{-7pt}
    \begin{subfigure}[b]{0.14\linewidth}
        \centering
        \includegraphics[width=\linewidth]{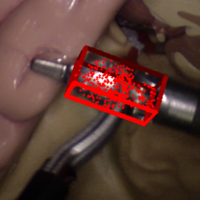}  
        \caption{}
    \end{subfigure}
    \hspace{-7pt}
    \begin{subfigure}[b]{0.14\linewidth}
        \centering
        \includegraphics[width=\linewidth]{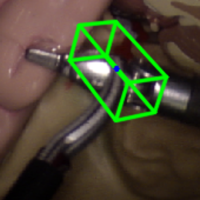}  
        \caption{}
    \end{subfigure}

    \caption{Qualitative results of zero-shot surgical tool pose estimation. The columns represent: (a) input scenes, (b) Mask R-CNN segmentation, (c) SAM segmentation, (d) SAM-6D (SAM) pose, (e) \textbf{Ours} SAM-6D (Mask R-CNN) pose, and (f) FoundationPose (Mask R-CNN mask) pose.}

    \label{fig:qual_res}
    \vspace{-10pt}
\end{figure*}

The tool segmentation validation is based on Dataset B, where the LND tool was partially occluded by other surgical instruments. The performance of Mask R-CNN was evaluated using two backbone configurations: ResNet-50 and ResNet-101. These results were compared against SAM. The results, as shown in Table \ref{tab:sam_segmentation_results}, indicate that Mask R-CNN with ResNet-101 consistently outperformed both ResNet-50 and SAM across all segmentation metrics, particularly in handling occlusions and reflections (Fig \ref{fig:qual_res} b \& c).

\begin{table}[ht]
\centering
\caption{Validation of segmentation models on Dataset B.
}
\label{tab:sam_segmentation_results}
\begin{tabular}{|c|cc|c|}
\hline
\multirow{2}{*}{\textbf{Metric}} & \multicolumn{2}{c|}{\textbf{Mask R-CNN}} & \multirow{2}{*}{\textbf{SAM}} \\ \cline{2-3}
 & \textbf{ResNet-50} & \textbf{ResNet-101} &  \\ \hline
\textbf{AP@[IoU=0.50:0.95]} & 85.7 & \textbf{86.9} & 46.4 \\ \hline
\textbf{AP$_s$ (Small)} & 21.8 & \textbf{22.2} & 21.6 \\ \hline
\textbf{AP$_m$ (Medium)} & 87.4 & \textbf{88.5} & 46.7 \\ \hline
\textbf{AP$_l$ (Large)} & 90.3 & \textbf{92.2} & 49.2 \\ \hline
\textbf{AR@[IoU=0.50:0.95]} & 87.4 & \textbf{88.7} & 54.4 \\ \hline
\end{tabular}
\vspace{-5pt}
\end{table}

\begin{table*}[ht]
\renewcommand{\arraystretch}{1.2}
\setlength{\tabcolsep}{2pt} 
\small
\resizebox{\textwidth}{!}{ 
\begin{tabular}{|c|ccccc|ccccc|ccccc|ccccc|}
\hline
\multirow{4}{*}{\textbf{Method}} & \multicolumn{10}{c|}{\textbf{Non-Occluded}} & \multicolumn{10}{c|}{\textbf{Occluded}} \\ \cline{2-21}
 & \multicolumn{5}{c|}{\textbf{ADD}} & \multicolumn{5}{c|}{\textbf{2D Projection}} & \multicolumn{5}{c|}{\textbf{ADD}} & \multicolumn{5}{c|}{\textbf{2D Projection}} \\ \cline{2-21}
 & \textbf{1mm} & \textbf{2.5mm} & \textbf{5mm} & \textbf{$\mu$} & \textbf{$\sigma$} & \textbf{5px} & \textbf{20px} & \textbf{50px} & \textbf{$\mu$} & \textbf{$\sigma$} & \textbf{1mm} & \textbf{2.5mm} & \textbf{5mm} & \textbf{$\mu$} & \textbf{$\sigma$} & \textbf{5px} & \textbf{20px} & \textbf{50px} & \textbf{$\mu$} & \textbf{$\sigma$} \\ \hline
FoundationPose & \textbf{8.86\%} & \textbf{43.14\%} & \textbf{48.29\%} & 8.05 & 17.45 & \textbf{37.14\%} & \textbf{48.86\%} & 90.29\% & 9.77e8 & 4.32e9 & 0.88\% & 4.77\% & 6.02\% & 84.90 & 46.86 & 1.38\% & 9.28\% & 28.73\% & 1.39e10 & 9.26e9 \\ \hline
\textbf{(Ours)}  & 3.14\% & 28.29\% & 46.86\% & \textbf{6.47} & \textbf{9.56} & 16.00\% & 31.43\% & \textbf{95.14\%} & \textbf{28.18} & \textbf{16.41} & \textbf{5.65\%} & \textbf{19.20\%} & \textbf{49.06\%} & \textbf{7.20} & \textbf{12.48} & \textbf{13.05\%} & \textbf{41.15\%} & 98.87\% & \textbf{21.88} & \textbf{12.93} \\ \hline
SAM-6D (SAM) & 0.00\% & 0.13\% & 0.88\% & 62.75 & 31.37 & 0.00\% & 0.13\% & 4.14\% & 5.05e7 & 1.01e9 & 0.00\% & 0.00\% & 0.86\% & 83.59 & 40.53 & 0.00\% & 0.00\% & 0.86\% & 3.45e8 & 2.61e9 \\ \hline
OVE-6D & 0.00\% & 0.86\% & 39.43\% & 14.73 & 37.93 & 0.00\% & 2.00\% & 92.57\% & 37.98 & 18.85 & 0.00\% & 0.00\% & 5.03\% & 100.95 & 48.07 & 0.00\% & 3.64\% & \textbf{99.37\%} & 165.02 & 1891.75 \\ \hline
MegaPose & 0.13\% & 3.01\% & 16.56\% & 8.42 & 11.69 & 3.76\% & 25.60\% & 93.98\% & 31.78 & 19.55 & 0.00\% & 0.39\% & 3.21\% & 16.79 & 23.81 & 0.10\% & 7.11\% & 51.31\% & 50.06 & 20.17 \\ \hline
\end{tabular}
}
\captionsetup{belowskip=-5pt}
\caption{Validation of pose estimation using ADD at 1mm, 2.5mm, 5mm thresholds, 2D Projection at 5px, 20px, and 50px thresholds, along with their corresponding means ($\mu$) and standard deviations ($\sigma$) (in mm and pixels for ADD and 2D Projection, respectively).}
\label{tab:pose_results_combined}
\vspace{-10pt}
\end{table*}

\begin{figure}[ht]
    \centering
    \includegraphics[width=0.49\linewidth]{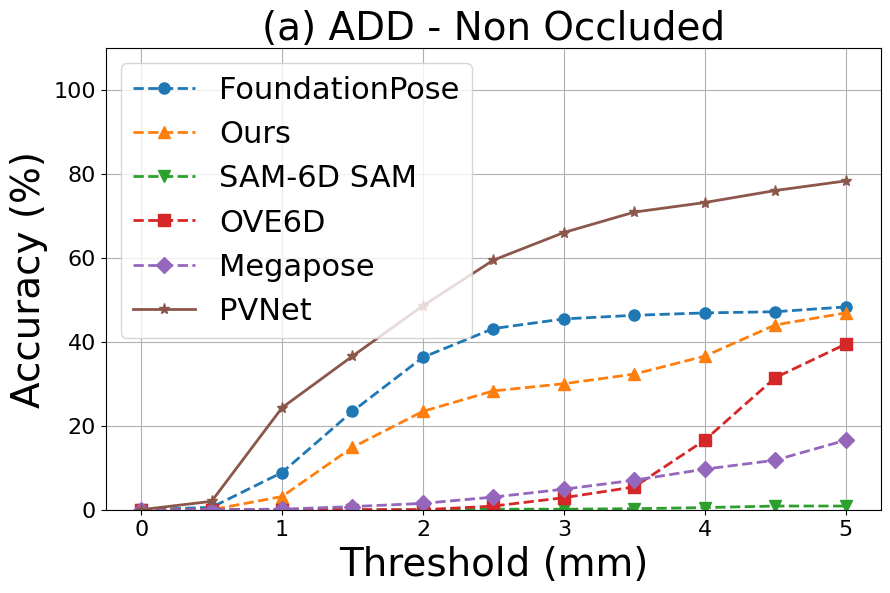}  
    \hfill
    \includegraphics[width=0.49\linewidth]{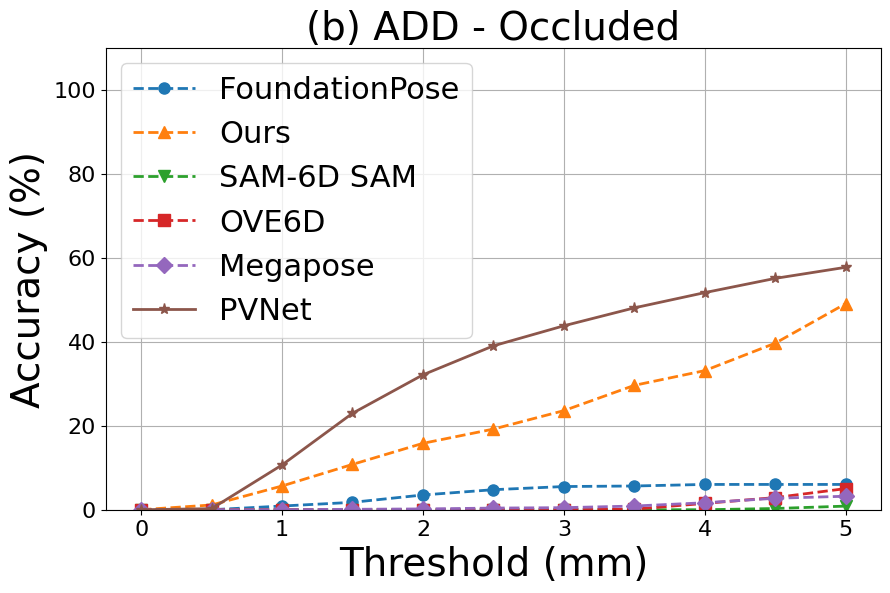}      

    \includegraphics[width=0.49\linewidth]{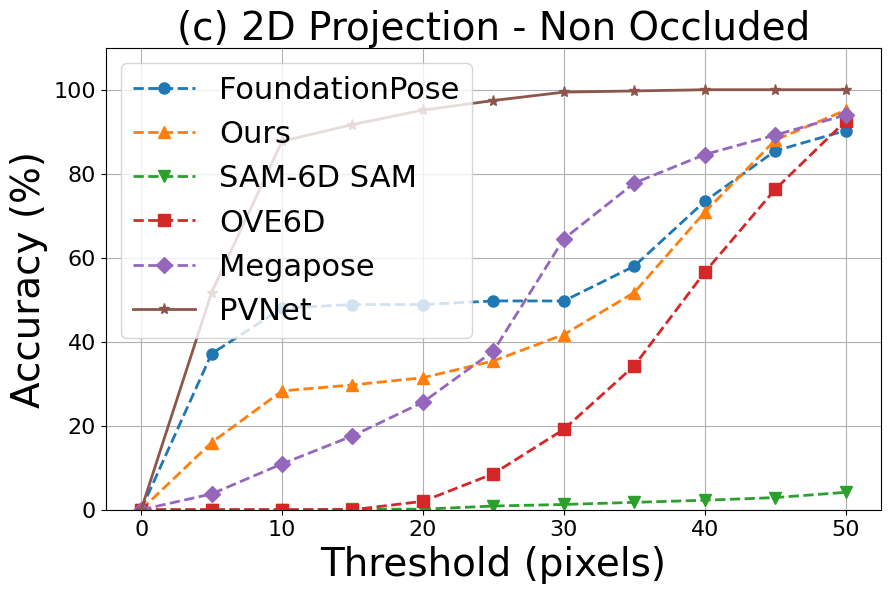}   
    \hfill
    \includegraphics[width=0.49\linewidth]{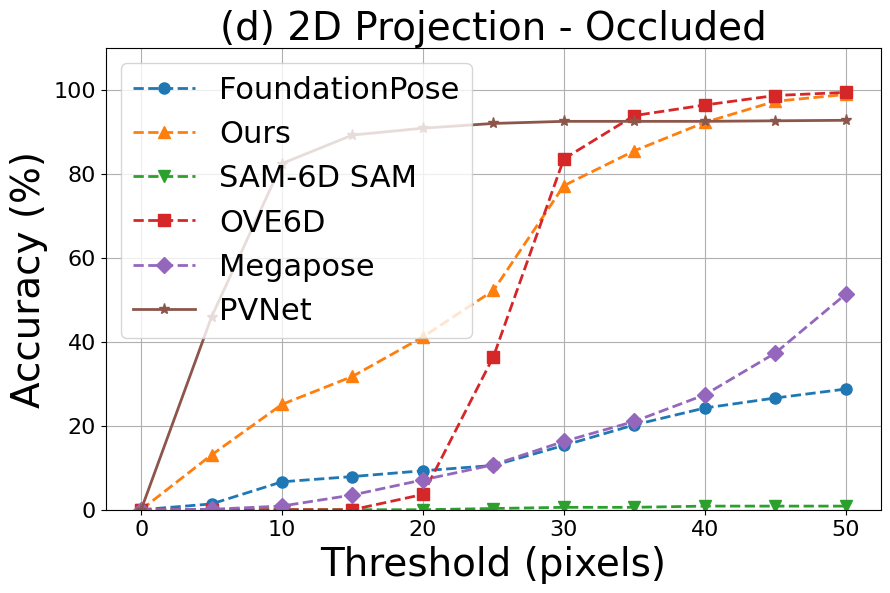}       
    \captionsetup{belowskip=-5pt}
    \caption{Validation of pose estimation models on Datasets A and B.
    (a, b) ADD and (c, d) 2D Projection accuracy.}
    \label{fig:comparison}
    \vspace{-10pt}
\end{figure}

\subsection{Validation of Zero-Shot Pose Estimation}

The performance evaluation of our zero-shot pose estimation pipeline using different models was performed in both non-occluded and occluded scenarios using the ADD and the 2D Projection metrics. The primary objective of this analysis was to assess the performance improvement brought by the replacement of SAM with the proposed fine-tuned Mask R-CNN in the SAM-6D framework, and to compare this enhanced model against state-of-the-art approaches such as FoundationPose, OVE-6D and MegaPose.

\subsubsection{Non-Occluded Scenario}
The results in Table \ref{tab:pose_results_combined} clearly demonstrate that FoundationPose achieves the best overall performance, leading in both ADD and 2D projection accuracy (Fig. \ref{fig:comparison} a \& c). For the ADD metric, FoundationPose achieves 48.29\% accuracy at the 5mm threshold, setting a strong benchmark. However, the proposed modification to SAM-6D with Mask R-CNN shows remarkable improvements, with an accuracy of 46.86\% at the 5mm threshold, narrowing the gap significantly. This is a substantial leap compared to the original SAM-6D (SAM), which only managed 0.88\% at the same threshold, underscoring the critical role of accurate mask generation in pose estimation. In terms of the 2D Projection metric, SAM-6D (Mask R-CNN) achieves even more compelling results. At the 50px threshold, SAM-6D (Mask R-CNN) achieves a superior accuracy of 95.14\%, exceeding FoundationPose, which records 90.29\%. 

\subsubsection{Occluded Scenario}

In the occluded scenario, the performance of all models naturally degrades due to the increased complexity of the task. Nonetheless, the proposed SAM-6D (Mask R-CNN) again demonstrates significant performance improvement over the original SAM-6D (SAM) model and even outperforms FoundationPose at all thresholds (Fig. \ref{fig:comparison} b \& d). At the 5mm ADD threshold, SAM-6D (Mask R-CNN) achieves 49.06\% accuracy, outperforming FoundationPose, which records just 6.02\%. Similarly, SAM-6D (Mask R-CNN) excels in the 2D projection task, achieving 98.87\% accuracy at the 50px threshold, compared to FoundationPose's 28.73\%. This performance improvement demonstrates the robustness of the proposed method in scenarios where partial visibility and occlusions are present—key challenges in surgical applications.

\subsubsection{Statistical Consistency and Robustness}

Table \ref{tab:pose_results_combined} further strengthens the case for our proposed enhancements. The mean \( \mu \) ADD for SAM-6D (Mask R-CNN) is 6.47mm, compared to 8.05mm for FoundationPose, indicating higher precision in pose estimation for the enhanced model. Additionally, the standard deviation \( \sigma \) of the ADD metric for SAM-6D (Mask R-CNN) is considerably lower (9.56mm) compared to FoundationPose (17.45mm). This reduction in variability suggests that the SAM-6D with Mask R-CNN delivers not only accurate but also consistent performance across the dataset, which is essential for the reliability of pose estimation in RMIS. The original SAM-6D (SAM), by contrast, demonstrates a far higher mean ADD of 62.75mm, with a large standard deviation of 31.37mm, reflecting its struggles in non-occluded environments. 

\subsubsection{Challenges with Other Models}

Both OVE-6D and MegaPose show comparatively weaker performance across both non-occluded and occluded scenarios. While OVE-6D manages decent performance in certain metrics, its overall mean ADD of 100.95mm in the occluded scenario indicates that it struggles significantly under occlusion. Similarly, MegaPose shows limited success, particularly in the occluded environment, where it records an accuracy of just 3.21\% at the 5mm ADD threshold.

The results clearly demonstrate that the proposed SAM-6D (Mask R-CNN) modification offers a substantial improvement over the original SAM-6D (SAM) (Fig \ref{fig:qual_res} d \& e) and performs competitively with FoundationPose in non-occluded scenarios and outperforms state-of-the-art in occluded environments as seen in Fig \ref{fig:qual_res} (e). Although supervised methods like PVNet \cite{peng2019pvnet} generally exhibit better performance (Fig. \ref{fig:comparison}), they require extensive training data, making them less flexible for dynamic surgical settings. In contrast, the results indicate that precise and consistent mask generation through Mask R-CNN is critical for accurate pose estimation, particularly in scenarios with occlusions and reflections. Regarding the computational time of SAM-6D (Mask R-CNN), the first frame of the video sequence requires 2.52s for the pose initialisation, while for subsequent frames, the model runs in near real-time with 15fps.


\section{CONCLUSIONS}

This work presents a novel zero-shot surgical instrument pose estimation pipeline for RMIS, combining stereo-based depth estimation with an enhanced instance segmentation model. By modifying SAM-6D with a fine-tuned Mask R-CNN, we achieved superior performance, particularly in occluded scenarios. Our approach outperforms state-of-the-art methods like FoundationPose in challenging environments, demonstrating its robustness and adaptability.
The use of synthetic data for fine-tuning Mask R-CNN enhances generalisability while reducing reliance on manual data collection. Our comprehensive evaluation against leading zero-shot models establishes a new benchmark in 6DOF pose estimation for RMIS, emphasising the importance of accurate mask generation and depth estimation.
This work sets a new standard for zero-shot models in surgical robotics, offering a generalisable solution that improves precision and adaptability. These advancements contribute to more efficient and reliable RMIS systems, potentially leading to improved surgical outcomes and patient safety.





\bibliographystyle{vancouver}
\bibliography{references}

\end{document}